\documentclass[final,1p,review,times]{elsarticle}

\usepackage{algorithm} 
\usepackage{algpseudocode} 
\usepackage{pdflscape}
\usepackage{adjustbox}
\usepackage[utf8]{inputenc}
\DeclareUnicodeCharacter{00B2}{\ensuremath{{}^2}}
\usepackage[edges]{forest}
\usetikzlibrary{arrows.meta}
\usetikzlibrary{calc}

\usepackage[figuresright]{rotating}
\usepackage{graphicx}
\usepackage{amssymb}
\usepackage{amsmath}

\usepackage[font=normalsize,labelfont=bf]{caption}
\usepackage{longtable}
\usepackage{natbib}

\usepackage{xcolor}

\usepackage{lineno}
\begin{document}

\begin{frontmatter}


\title{Ensemble deep learning: A review}



 \author[label1]{M.A. Ganaie}
 \ead{phd1901141006@iiti.ac.in}
 \author[label2]{Minghui Hu}
\ead{minghui.hu@ntu.edu.sg}
 \author[label1]{A.K. Malik}
 \ead{phd1801241003@iiti.ac.in}
\author[label1]{M. Tanveer\corref{cor1}}
\ead{mtanveer@iiti.ac.in}
\cortext[cor1]{Corresponding authors}
\address[label1]{Department of Mathematics, Indian Institute of Technology Indore, Simrol, Indore, 453552, India}
\address[label2]{School of Electrical \& Electronic Engineering, Nanyang Technological University, Singapore\fnref{label4}}
\address[label3]{KINDI Center for Computing Research College of Engineering, Qatar University, Qatar}





\author[label2,label3]{P.N. Suganthan\corref{cor1}}
\ead{epnsugan@ntu.edu.sg}

\begin{abstract}
Ensemble learning combines several individual models to obtain better generalization performance. 
Currently, deep learning  architectures are showing better performance compared to the shallow or traditional models.
 Deep ensemble learning models combine the advantages of both the deep learning models as well as the ensemble learning such that the final model has better generalization performance. 
 This paper reviews the state-of-art deep ensemble models and hence serves as an extensive summary for the researchers.
 The ensemble models are broadly categorised into  bagging, boosting,  stacking, negative correlation based deep ensemble models, explicit/implicit ensembles, homogeneous/heterogeneous ensemble, decision fusion strategies based deep ensemble models. Applications of deep ensemble models in different  domains are also briefly discussed. Finally, we conclude this paper with some potential future research directions. 
\end{abstract}


\end{frontmatter}

\section{Introduction}
Deep learning architectures have been successfully employed across a wide range of applications from image/video classification to the health care. The success of these models is attributed to the better feature representation via multi layer processing architectures. The deep learning models have been mainly used for classification, regression and clustering problems.
Classification problem is defined as the categorization of the new observations based on the hypothesis $h$ learned from the set of training data. The hypothesis $h$ represents a mapping of input data features to the appropriate target labels/classes. The main objective, while learning the hypothesis $h$, is that it should approximate the true unknown function as close as possible to reduce the generalization error. There exist several applications of these classification algorithms ranging from medical diagnosis to remote sensing. Mathematically, 
\begin{align}
    O_c=h(x,\theta_c), \hspace{1mm} O_c\in \mathbb{Z},
\end{align}
where $x$ is the input feature vector, $O_c$ is the category of the sample $x$, $\theta_c$ is the set of learning parameters of the hypothesis $h$ and $\mathbb{Z}$ is the set of class labels. 

Regression problems deal with the continuous decisions, instead of discrete categories. Mathematically,
\begin{align}
     O_r=h(x,\theta_r), \hspace{1mm} O_r\in \mathbb{R},
\end{align}
where $x$ is the observation vector, $O_r$ is the output, and $\theta_r$ is the set of learning parameters of the hypothesis $h$. 

Broadly speaking, there are different approaches of classification like  supervised, unsupervised classification, few-shot, one-shot and so on. Here, we only discuss supervised and unsupervised classification problems. In supervised learning, the building of hypothesis $h$ is supervised based on the known output labels provided in the training data samples, while as in unsupervised learning hypothesis $h$ is generated without any supervision as no known output values are available with the training data. This approach, also known as clustering, generates the hypothesis $h$  based on the similarities and dissimilarities present in the training data.

Generally speaking, the goal of generating the hypothesis $h$ in Machine learning area is that it should perform better when applied to unknown data. The performance of the model is measured with respect to the area in which the model is applied. Combining the predictions from several models has proven to be an elegant approach for increasing the performance of the models. Combination of several different predictions from different models to make the final prediction is known as ensemble learning or ensemble model. 
The ensemble learning involves multiple  models combined in some fashion like averaging, voting such that the ensemble model is better than any of the individual models. To prove that average voting in an ensemble is better than individual model, Marquis de Condorcet proposed a theorem wherein he proved that if the probability of each voter being correct is above $0.5$ and the voters are independent, then addition of more voters increases the probability of majority vote being correct until it approaches $1$ \cite{condorcet1785essay}. Although Marquis de Condorcet proposed this theorem in the field of political science and had no idea of the field of Machine learning, but it is the similar mechanism that leads to better performance of the ensemble models. Assumptions of Marquis de Condorcet theorem also holds true for ensembles \cite{hansen1990neural}.
The reasons for the success of ensemble learning include: statistical, computational and representation learning \cite{dietterich2000ensemble}, bias-variance decomposition \cite{kohavi1996bias} and strength-correlation \cite{breiman2001random}.

In this era of machine learning, deep learning automates the extraction of high-level features via hierarchical feature learning mechanism wherein  the upper layer of features are generated on the previous set of layer/layers.  Deep learning has been successfully applied across different fields  since the ImageNet Large Scale Recognition Challenge (ILSVRC) competitions \cite{russakovsky2015imagenet, krizhevsky2012imagenet} and has achieved state-of-art performance. It has obtained promising results in object detection, semantic segmentation, edge detection and number of other domains. However, given the computational cost, the training of deep ensemble models is an uphill task. Different views have been provided to understand how the deep learning models learn the features like learning through hierarchy of concepts via many levels of representation \cite{deng2014deep, goodfellow2016deep, lecun2015deep}. 
Given the advantages of deep learning models from deep architectures, there are several bottlenecks like vanishing/exploding gradients \cite{hochreiter1991untersuchungen, glorot2010understanding} and degradation problem \cite{he2016deep} which prevent to reach this goal. Recently, training deep network's has become feasible through the Highway networks \cite{srivastava2015training} and Residual networks \cite{he2016deep}. Both these networks enabled to train very deep networks. The ensemble learning has been recently known to be strong reason for enhancing the performance of deep learning models \cite{veit2016residual}. Thus, the objective of deep ensemble models is to obtain a model that has best of both the ensemble and deep models. 

There exist multiple surveys in the literature which mainly focus on the review of ensemble learning like learning of ensemble models in classification problems \cite{zhao2005survey, rokach2010ensemble, gopika2014analysis, yang2010review}, regression problems \cite{ mendes2012ensemble, ren2015ensemble} and clustering \cite{vega2011survey}. Review of both the classification and regression models was given in \cite{ren2016ensemble}. Comprehensive review of the ensemble methods and the challenges were given in \cite{sagi2018ensemble}. Though \cite{sagi2018ensemble} provided some insight about the deep ensemble models but couldn't give the comprehensive review of the deep ensemble learning while as \cite{cao2020ensemble} reviewed the ensemble deep models in the context of bioinformatics. 
The past decade has successively evolved different deep learning strategies which have lead to the exploration and innovation of these models in multiple areas like health care, speech, image classification, forecasting and other applications. Broadly speaking, ensemble learning approaches have followed classical methods, general methods and different fusion strategies for improving the performance of the models. Since deep learning models are computation and data extensive, hence, ensemble deep learning models need special attention while exploring the complementary information of multiple algorithms into a uniform framework. Ensemble deep learning models need to handle multiple questions like how to induce diversity among the baseline models, how to keep the training time as well the models complexity lower for the practical applications, how to fuse the predictions of the complementary algorithms. Multiple studies have handled these problems differently. In this review paper, we comprehensively review the different approaches  used to handle the aforementioned problems.
In this paper, we give a comprehensive review of deep ensemble models. {\bf {To the best of our knowledge, this is the first comprehensive review paper on deep ensemble models.}}

The rest of this paper is organised as follows: Section-\ref{sec:theory} discusses the theoretical aspects of deep ensemble learning, Section-\ref{sec:Ensemble Strategies:} discusses the different approaches used in deep ensemble strategies, applications  of deep ensemble methods are given in Section-\ref{sec:Applications} and finally conclusions and future directions are given in Section-\ref{sec:Conclusions and future works}.




\begin{landscape}
 \begin{figure}
	\centering
	\begin{adjustbox}{max size={\paperheight}{1\textheight}}
\includegraphics[]{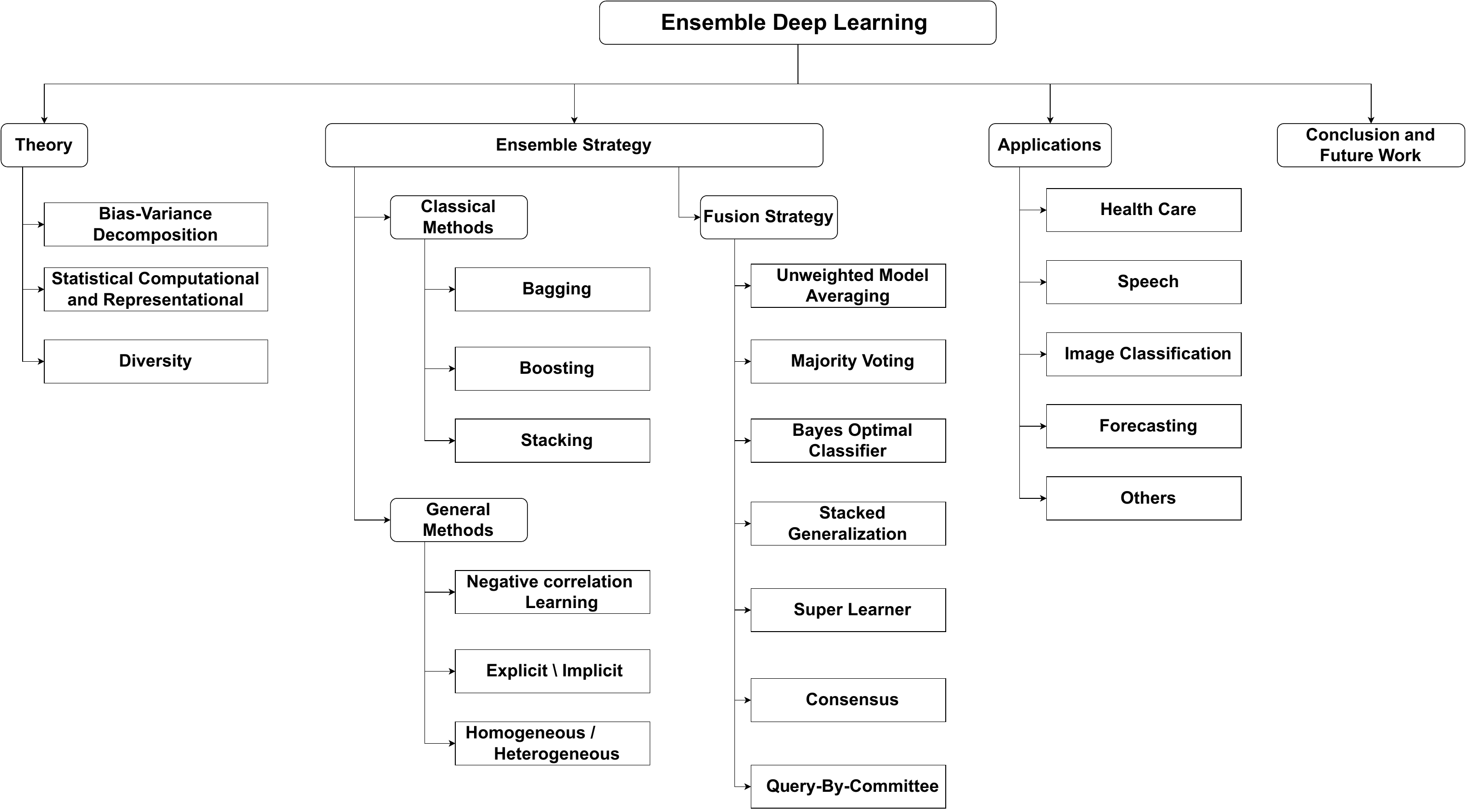}
\end{adjustbox}
\caption{Layout of the paper.} \label{fig:connect_paper}
\end{figure}
\end{landscape}


\section{Research Methodology}
The studies in this review are searched from the  Google Scholar and Scopus search engines. The papers are the result of ensemble learning, ensemble deep learning, deep ensemble learning, deep ensembles keywords. The articles where screened based on the title and abstract, followed by the screening of full-text version. The articles are elaborated based on the ensemble learning and deep learning approaches.  

\section{Theory}
\label{sec:theory}
The various reasons which have been justified for the success of ensemble learning can be discussed under the following subheadings:

\subsection{\textbf{Bias-Variance Decomposition}} Initially, the success of ensemble methods was theoretically investigated for regression problems. \citet{krogh1995neural, brown2005diversity} proved via ambiguity decomposition  that the proper ensemble classifier guarantees a smaller squared error as compared to the individual predictors of the classifier. Ambiguity decomposition was given for single dataset based ensemble methods, later on, multiple dataset bias-variance-covariance decomposition was introduced in \cite{brown2005diversity, geman1992neural, brown2005managing, pedro2000unified}  and is given as:
\begin{align}
    E[o-t]^2&=bias^2+\frac{1}{M}var+(1-\frac{1}{M})covar, \nonumber\\
    bias&=\frac{1}{M}\sum_i(E[o_i]-t), \nonumber\\
    var&=\frac{1}{M}\sum_iE[o_i-E[o_i]]^2,\\
    covar&=\frac{1}{M(M-1)}\sum_i\sum_{j\neq i}E[o_i-E[o_i]][o_j-E[o_j]], \nonumber
\end{align}
where $t$ is target, $o_i$ is the output of $i^{th}$ model and $M$ is the ensemble size. Here, $bias$ term measures the average difference between the base learner and the model output, $var$ indicates their average variance, and $covar$ is the covariance term measuring the pairwise difference of the base learners. 

Ensemble methods have been supported by several theories like bias-variance \cite{kohavi1996bias, wolpert1997bias}, strength correlation \cite{breiman2001random}, stochastic discrimination \cite{kleinberg1990stochastic}, and margin theory \cite{schapire1998boosting}. These theories provide the equivalent of bias-variance-covariance decomposition \cite{pisetta2012new}.  

The above given equations of decomposition error can't be directly applied to the datasets with discrete class labels due to their categorical nature. However, alternate ways to decompose the error in classification problems are given in \cite{kohavi1996bias, kong1995error, friedman1997bias, breiman1998arcing, james2003variance}.

Multiple approaches  like bagging, boosting have been proposed for generating the ensemble methods. Bagging reduces the variance among the base classifiers \cite{breiman1996bagging}  while as boosting based ensembles lead to the bias and variance reduction \cite{breiman1996bias, zhang2008rotboost}. 

\subsection{\textbf{Statistical, Computational and Representational Aspects}}
Dietterich provided Statistical, Computational and Representational reasons \cite{dietterich2000ensemble} for success of ensemble models. The learning model is viewed as the search of the optimal hypothesis $h$ among the several hypothesis in the search space. When the amount of data available for the training is smaller compared to the size of the hypothesis space, the statistical problem arises. Due to this statistical problem, the learning algorithm identifies the different hypothesis which gives same performance on the training samples. Ensembling of these hypothesis results in an algorithm which reduces the risk of being a wrong classifier. The second reason is computational wherein a learning algorithm stucks in a local optima due to some form of local search. Ensemble model overcomes this issue by performing some form of local search via different starting points   which leads to better approximation of the true unknown function.
Another reason is representational wherein none of the hypotheses among the set of hypothesis is able to represent the true unknown function. Hence, ensembling of these hypothesis via some weighting technique results into the hypothesis which expands  the representable function space.

\subsection{\textbf{Diversity}}
One of the main reasons behind the success of ensemble methods is increasing the diversity among the base classifiers and the same thing was highlighted in \cite{dietterich2000ensemble}. Different approaches have been followed to generate diverse classifiers. Different methods like bootstrap aggregation (bagging) \cite{breiman1996bagging}, Adaptive Boosting (AdaBoost) \cite{freund1996experiments}, random subspace \cite{barandiaran1998random}, and random forest \cite{breiman2001random} approaches are followed for generating the multiple datasets from the original dataset to train the different predictors such that the outputs of predictors are diverse. Attempts have been made to increase diversity in the output data  wherein multiple outputs are created instead of multiple datasets for the supervision of the base learners. `Output smearing' \cite{breiman2000randomizing} is one of this kind which induces random noise to introduce diversity in the output space.



\section{Ensemble Strategies: }
\label{sec:Ensemble Strategies:}
The different ensemble strategies have evolved over a period of time which results in better generalization of the learning models. The ensemble strategies are broadly categorised as follows: 

\begin{figure}
    \centering
    \includegraphics[width=0.3\textwidth]{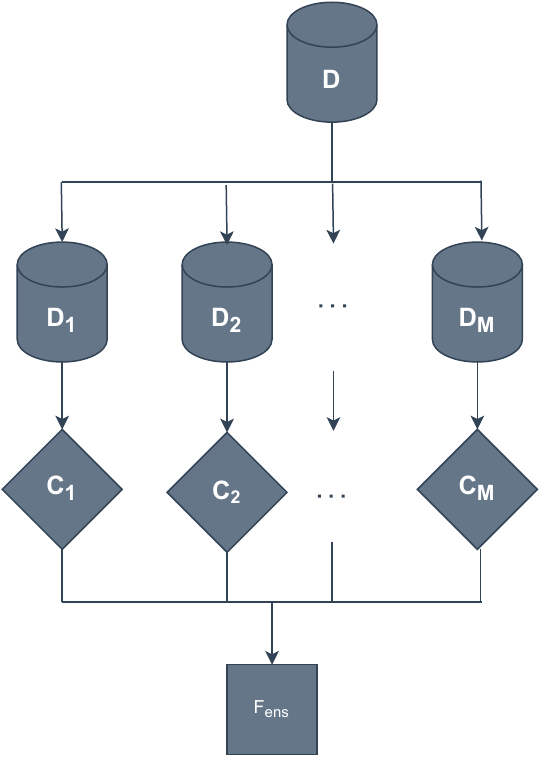}
    \caption{Bagging}
    \label{Fig:Bagging}
    
    \includegraphics[width=0.3\textwidth]{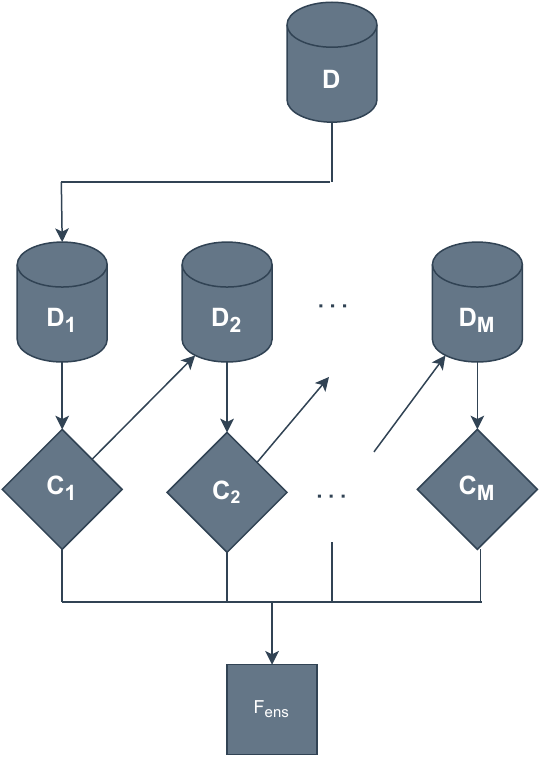}
    \caption{Boosting}
    \label{Fig:Boosting}
    
    \includegraphics[width=0.3\textwidth]{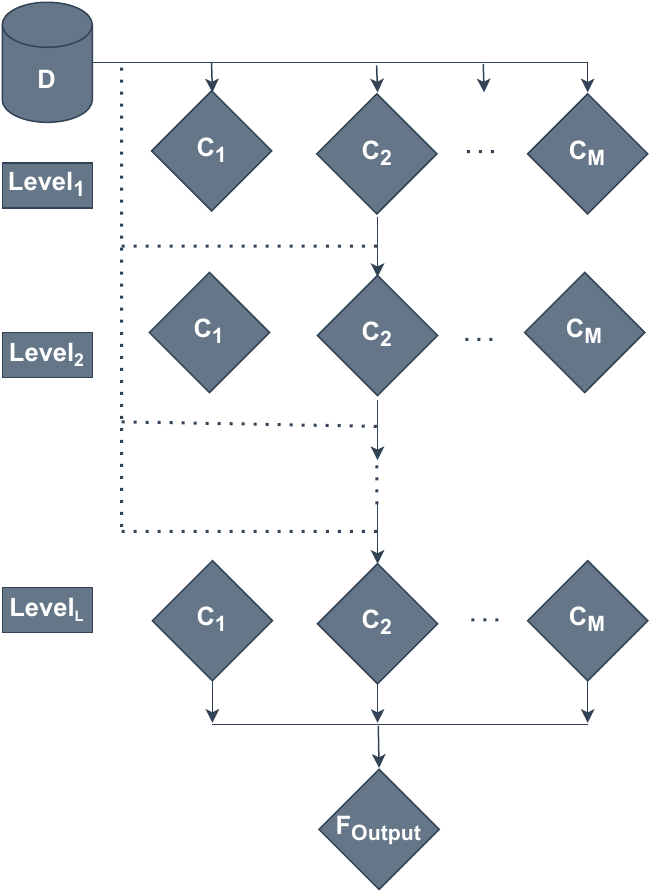}
    \caption{Stacking}
    \label{Fig:Stacking}
\end{figure}

\subsection{\textbf{Bagging}}
Bagging \cite{breiman1996bagging}, also known as bootstrap aggregating, is one of the standard techniques for generating the  ensemble-based algorithms. Bagging is applied to enhance the performance of an ensemble classifier. 
The main idea in bagging is to generate a series of independent observations with the same size, and distribution  as that of the original data.  Given the series of observations, generate an ensemble predictor which is better than the single predictor generated on the original data.
Bagging increases two steps in the original models: First, generating the bagging samples and passing each bag of samples to the base models and second, strategy for combining the predictions of the multiple predictors. 
Bagging samples may be generated with or without replacement. Combining the output of the base predictors may vary as mostly majority voting is used for classification problems while the averaging strategy is used in regression problems for generating the  ensemble output. Figure \ref{Fig:Bagging} shows the diagram of the bagging technique. Here, $D_{i}$ represents the bagged datasets, $C_i$ represents the algorithms and $F_{ens}$ calculates the final outcome.

Random Forest \cite{breiman2001random} is an improved version of the decision trees that uses the bagging strategy for improving the predictions of the base classifier which is a decision tree. The fundamental difference between these two methods is that at each tree split in Random Forest, only a subset of features is randomly selected and considered for splitting. The purpose of this method is to decorrelate the trees and prevent over-fitting. \citet{breiman2001random} showed heuristically that the variance of the bagged predictor is smaller than the original predictor and proposed that bagging is better in higher dimensional data.  However, the analysis of the smoothing effect of bagging \cite{buja2000smoothing} revealed that bagging doesn't depend on the data dimensionality.  

\citet{buhlmann2002analyzing} gave theoretical explanation of how bagging gives smooth hard decisions, small variance, and mean squared error. Since bagging is computationally expensive, hence subbagging and half subbagging \cite{buhlmann2002analyzing} were introduced. Half subbagging, being computationally efficient, is as accurate as the bagging.

Several attempts tried to combine bagging with other machine learning algorithms. \citet{kim2002support} used bagging method to generate multiple bags of the dataset and multiple support vector machines were trained independently with each bag as the input. The output of the models is  combined via majority voting, least squares estimation weighting and double layer hierarchical approach. In the double layer hierarchical approach, another support vector machines (SVM) is used to combine the outcomes of the multiple SVM's efficiently. \citet{tao2006asymmetric} used asymmetric bagging strategy to generate the ensemble model to handle the class imbalance problems. 
A case study of bagging, boosting and basic ensembles \cite{mao1998case} revealed that at higher rejection rates of samples boosting is better as compared to bagging and basic ensembles. However, as the rejection rate increases the difference disappears among the boosting, bagging and basic ensembles. Bagging based multilayer perceptron \cite{ha2005response} combined bagging to train multiple  perceptrons with the corresponding bag and showed that bagging based ensemble models perform better as compared to individual multilayer perceptron. In \cite{genccay2001pricing}, the analysis  of the bagging approach and other regularisation techniques revealed that bagging regularized the neural networks and hence provide better generalization. In \cite{khwaja2015improved}, bagged neural networks (BNNs) was proposed wherein each neural network was trained over different dataset sampled randomly with replacement from original dataset and was implemented for  the short term load forecasting.
 Unlike Random forest \cite{breiman2001random} which uses majority voting for aggregating the ensemble of decision trees, bagging based survival trees \cite{hothorn2004bagging} used Kaplan–Meier curve to predict the ensemble output for breast cancer and lymphoma patients.
 In \cite{alvear2018building}, ensembles of stacked denoising autoencoders for classification showed that the bagging and switching technique in a general deep machine results in improved diversity. 

Bagging has also been applied to solve the problem of imbalanced data. Roughly Balanced Bagging \cite{hido2009roughly} tries to equalize each class's sampling probability in binary class problems wherein the negative class samples are sampled via negative binomial distribution, instead of keeping the sample size of each class the same number. Neighbourhood Balanced Bagging \cite{blaszczynski2015neighbourhood} incorporated the neighbourhood information for generating the bagging samples for the class imbalance problems. \citet{blaszczynski2015neighbourhood} concluded that applying conventional diversification is more effective when applied at the last classification methods. Both roughly balanced Bagging and Neighbourhood Balanced Bagging have not been explored  in deep learning architectures. Thus, these approaches can be exploited to handle the class imbalance problems via deep ensemble models.

The theoretical and experimental analysis of online bagging and boosting \cite{oza2005online} showed that the online bagging algorithm can achieve similar accuracy as the batch bagging algorithm with only a little more training time. However, online bagging is an option when all training samples can't be loaded into the memory due to memory issues. 

Although ensembling may lead to increase in the computational complexity, but bagging possesses the property that it can be paralleled and can lead to effective reduction in the training time subject to the availability of hardware for running the parallel models. Since deep learning models have high training time, hence optimization of multiple deep models on different training bags is not a feasible option. 

\begin{table}[]
    \centering
    \resizebox{\textwidth}{!}{
    \begin{tabular}{|l|l|l|}
    \hline
    Years&Authors&Contribution\\
    \hline
   1996& \citet{breiman1996bagging}& Proposed the idea of Bagging\\
    1998&  \citet{mao1998case} & Case study of bagging, boosting and basic ensembles\\
   2000&  \citet{buja2000smoothing}& Theoretical analysis of bagging\\
      2001&  \citet{breiman2001random}  & Bagging with random subspace Decision trees and ensembling  outputs via majority voting \\
       2001&  \citet{genccay2001pricing}& Study of Bayesian regularization, early stopping and Bagging \\
        2002&   \citet{kim2002support}& Bagging with SVM's and ensembling  outputs via SVM's, majority voting and least squares estimation\\
      2002&   \citet{buhlmann2002analyzing}& Theoretical justification of Bagging, proposed subbagging and half subagging\\
      2004& \citet{hothorn2004bagging}& Bagging with decision trees and ensembling  outputs via Kaplan–Meier curve \\
      
       2005&   \citet{oza2005online} & Theoretical and experimental analysis of online bagging and boosting\\
      2006&   \citet{tao2006asymmetric}& Proposed assymmetric bagging with SVM's and ensembling  outputs SVM's \\
       2009&   \citet{hido2009roughly}& Roughly balanced bagging on decision trees and ensembling  outputs via majority voting\\
      2005, 2015&    \citet{ha2005response,khwaja2015improved} & Bagging with Neural networks and ensembling  outputs via majority voting\\
    2015&   \citet{blaszczynski2015neighbourhood}& Neighbourhood balanced bagging ensembling  outputs via majority voting\\
     \hline
    \end{tabular}}
    \caption{Bagging based ensemble models} 
    \label{tab:Bagging}
\end{table}
\subsection{\textbf{Boosting}} 
Boosting technique is used in ensemble models for converting a weak learning model into a learning model  with better generalization.  Figure \ref{Fig:Boosting} shows the diagram of the boosting technique.  The techniques such as majority voting in case of classification problems or a linear combination of weak learners in the regression problems results in better prediction as compared to the single weak learner. Boosting methods like AdaBoost \cite{freund1996experiments} and Gradient Boosting \cite{friedman2001greedy} have been used across different domains. Adaboost uses a greedy technique for minimizing a convex surrogate function upper bounded by misclassification loss via augmentation, at each iteration, the current model with the appropriately weighted predictor. AdaBoost learns an effective ensemble classifier as it leverages the incorrectly classified sample at each stage of the learning. AdaBoost minimizes the exponential loss function while as the Gradient boosting generalized this framework to the arbitrary differential loss function. 

Boosting, also known as forward stagewise additive modelling, was  originally proposed to improve the performance of the classification trees. 
It has been recently incorporated in the deep learning models to further improve their performance. 

Boosted deep belief network (DBN) \cite{liu2014facial} for facial expression recognition unified the boosting technique and multiple  DBN's via objective function which results in a strong classifier. The model learns complex feature representation to build a strong classifier in an iterative manner.
Deep boosting \cite{Cortes2014} is an ensemble model that uses the deep decision trees. It can also  be used in combination with any other rich family classifier   to improve the generalization performance. In each stage of the deep boosting, the decisions of which classifier to add and what weights should be chosen depends on the (data-dependent) complexity of the classifier to which it belongs. The interpretation of the deep boosting classifier is given via structural risk minimization principle  at each stage of the learning. Multiclass Deep boosting \cite{Kuznetsov2014} extended the  Deep boosting \cite{Cortes2014} algorithm to theoretical, algorithmic, and empirical results to the multiclass problems. Due to the limitation of the training data in each mini batch, Boosting CNN may overfit the data. To avoid overfitting, incremental Boosting CNN (IBCNN) \cite{han2017incremental} accumulated the information of multiple batches of the  training data samples. IBCNN uses decision stumps on the top of single neurons as the weak learners and learns weights via AdaBoost method in each mini batch.  
Unlike DBN \cite{liu2014facial} which uses image patch for learning the weak classifiers,  IBCNN trains the weak classifiers from the fully connected layer i.e. the whole image is used for learning the weak classifiers. To make the IBCNN model more efficient, the weak learners loss functions are combined with the global loss function.

Boosted CNN \cite{moghimi2016boosted} used boosting for training the deep CNN. Instead of averaging, least squares objective function was used to incorporate the boosting weights into CNN. \citet{moghimi2016boosted} also showed that CNN can be replaced by network structure within their boosting framework for improving the performance of the base classifier. Boosting increases the complexity of training the networks, hence the concept of dense connections was introduced in a deep boosting framework to overcome the problem of vanishing gradient problem for image denoising \cite{chen2018deep}. Deep boosting framework was extended to image restoration in \cite{chen2019real} wherein the dilated dense fusion network was used to boost the performance.  

The convolutional channel features \cite{yang2015convolutional} generated the high level features via CNN and then used boosted forest for final classification. Since CNN has high number of hyperparameters than the boosted forest, hence the model proved to be efficient than end-to-end training of CNN models both in terms of performance and time. \citet{yang2015convolutional} showed its application in edge detection, object proposal generation, pedestrian and face detection.
 A stagewise boosting deep CNN \cite{walach2016learning} trains several models of the CNNs within the offline paradigm boosting framework. To extend the concept of boosting in online scenario's wherein only a chunk of data is available at  given time, Boosting Independent Embeddings Robustly (BIER) \cite{opitz2017bier} was proposed to cope up the online scenario's. In BIER, a single CNN model is trained end-to-end with an online boosting technique. The training set in the BIER is reweighed via the negative gradient of the loss function to project the input spaces (images) into a collection of independent output spaces. To make BIER more robust, Hierarchical Boosted deep metric learning \cite{waltner2019hibster} incorporated the hierarchical label information into the embedding ensemble which improves the performance of the model on the large scale image retrieval application. Using deep boosting results in higher training time, to reduce the warm-up phase of training which trains the classifier from scratch deep incremental boosting \cite{mosca2017deep} used transfer learning approach. This approach leveraged the initial warm-up phase of each incremental base model of the ensemble during the training of the network. To reduce the training time of boosting based ensembles, snapshot boosting \cite{zhang2020snapshot} combined the merits of snapshot ensembling and boosting to improve the generalization without increasing the cost of training. Snapshot boosting trains each base network and combines the outputs via meta learner to combine the output of base learners more efficiently.

 

Literature shows that the boosting concept is the backbone behind well-known architectures like Deep Residual networks \cite{he2016deep, Siu}, AdaNet \cite{Cortes2017} . The theoretical background for the success of the Deep Residual networks (DeepResNet) \cite{he2016deep} was explained in the context of boosting theory \cite{Huang2018}. 
The authors proposed multi-channel telescoping sum boosting  learning framework, known as BoostResNet, wherein each channel is a scalar value updated during rounds of boosting to minimize the multi-class error rate. The fundamental difference between the AdaNet and BoostResnet is that the former maps the feature vectors to classifier space and boosts weak classifiers while the latter used multi-channel representation boosting. Moreover, BoostResNet is more efficient than DeepResnet  in terms of computational time.

The theory of boosting was extended to online boosting in \cite{Beygelzimer} and provided theoretical convergence guarantees. Online boosting shows improved convergence guarantees for batch boosting algorithms.

The ensembles of bagging and boosting have been evaluated in \cite{gonzalez2020practical}. The study evaluated  the different algorithms based on the concept of bagging and boosting along with the availability of software tools. The study highlighted the practical issues and opportunities of their feasibility in  ensemble modeling. 

\begin{table}[]
    \centering
    \resizebox{\textwidth}{!}{
    \begin{tabular}{|c|c|l|}
    \hline
   Years&Authors& Contribution\\
    \hline
  2014&  \citet{liu2014facial}&Boosted deep belief network (DBN)  as base classifiers for facial expression recognition.\\
   2014&   \citet{Cortes2014}& Decision trees as base classifiers for binary class classification problems. \\
    2014&  \citet{Kuznetsov2014}& Decision trees as base classifiers for multiclass classification problems.\\
     2015&   \citet{yang2015convolutional}& Ensemble of CNN and boosted forest for  edge detection, object proposal generation, pedestrian and face detection.\\
     2016&  \citet{moghimi2016boosted}& Boosted CNN \\
     2016&\citet{walach2016learning}& CNN Boosting applied to bacterila cell images and crowd counting.\\
2017&\citet{opitz2017bier} & Boosted deep independent embedding model for online scenarios.\\
2017&\citet{mosca2017deep}& Transfer learning based deep incremental boosting.\\
    2017&  \citet{han2017incremental}&Boosting based CNN with incremental approach for facial action unit recognition.\\
    2018&   \citet{chen2018deep} & Deep boosting for image denoising with dense connections. \\
   2019&   \citet{chen2019real}& Deep boosting for image restoration and image denoising. \\
  2019&\citet{waltner2019hibster}&Hierarchical boosted deep metric learning with hierarchical label embedding.  \\
2020&\citet{zhang2020snapshot}&Snapshot boosting.\\
      \hline
    \end{tabular}}
    \caption{Boosting based ensemble models}
    \label{tab:Boosting}
\end{table}

\subsection{\textbf{Stacking}}
Ensembling can be done either by combining outputs of multiple base models in some fashion or using some method to choose the ``best" base model. Figure \ref{Fig:Stacking} shows the stacking technique. Stacking is one of the integration techniques wherein the meta-learning model is used to integrate the output of base models. If the final decision part is a linear model, the staking is often referred to as “model blending” or simply “blending”.
The concept of stacking or stacked regression was initially given by \cite{wolpert1992stacked}. In this technique, the dataset is randomly split into $J$ equal parts. For the $j^{th}$-fold cross-validation one set is used for testing and the rest are used for training. With these training testing pair subsets, we obtain the predictions of different learning models which are used as the meta-data to build the meta-model. Meta-model makes the final prediction, which is also called the winner-takes-all strategy. 

Stacking is a bias reducing technique \cite{Leblanc1996}. Following \cite{wolpert1992stacked},  Deep convex net (DCN) \cite{Deng2011} was proposed which is a deep learning architecture composed of a variable number of modules stacked together to form the deep architecture. Each learning module in DCN is convex.  DCN  is a  stack of several modules consisting of linear input units, hidden layer non-linear units, and the second linear layer with the number of units as that of target classification classes. The modules are connected layerwise as the output of the lower module is given as input to the adjacent higher module in addition to the original input data. The deep stacking network (DSN)  enabling parallel training on very large scale datasets was proposed in \cite{Deng2012}, the network was named stacking based as it shared the concept of ``stacked generalization" \cite{wolpert1992stacked}. The kernelized version of DCN, known as kernel deep convex networks (K-DCN), was given in \cite{Deng2012c},  here the number of hidden layer approach infinity via kernel trick. \citet{Deng2012c} showed that K-DCN performs better as compared to the DCN. However, due to kernel trick the memory requirements increase and hence may not be scalable to large scale datasets. Also, we need to optimize the hyperparameters like the number of levels in the stacked network, the kernel parameters to get the optimal performance of the network.  To leverage the memory requirements, random Fourier feature-based kernel deep convex network \cite{Huang2013b} approximated the Gaussian kernel which reduces the training time and helps in the evaluation of K-DCN over large scale datasets. A framework for parameter estimation and model selection in kernel deep stacking networks \cite{Welchowski2016b} is based on the combination of model-based optimization and hill-climbing approaches. \citet{Welchowski2016b} used data-driven framework for parameter estimation, hyperparameter tuning and model selection in kernel deep stacking networks. 
Another improvement over DSN was Tensor Deep Stacking Network (T-DSN) \cite{Hutchinson2012a}, here in each block of the stacked network, large single hidden layer was split into two smaller ones and then mapped bilinearly to capture the higher-order interactions among the features. Comprehensive evaluation, the more detailed analysis of the learning algorithm and T-DSN implementation is given in \cite{Hutchinson2013}. Sparse coding is another popular method that is used in the deep learning area. The advantage of sparse representation is numerous, including robust to noise, effective for learning useful features, etc. Sparse Deep Stacking Network (S-DSN) is proposed for image classification and abnormal detection \cite{li2015sparse, sun2018sparse}. \citet{li2015sparse, sun2018sparse} stacked many sparse simplified neural network modules (SNNM) with mixed-norm regularization, in which weights are solved by using the convex optimization and the gradient descent algorithm. In order to make sparse  SNNM learning the local dependencies between hidden units, \citet{li2017visual} split the hidden units or representations into different groups, which is termed as group sparse DSN (GS-DSN). The DSN idea is also utilized in the Deep Reinforcement Learning field. \citet{zhang2020grasp} employed DSN method to integrate the observations from the formal network: Grasp network and Stacking network based on Q-learning algorithm to make an integrated robotic arm system do grasp and place actions. 
\citet{wang2020particle} stacked blocks multiple times to increase the performance of the neural architecture search task. \citet{zhang2019deep} presents a deep hierarchical multi-patch network for image deblurring via stacking approach.

Since there is no temporal representation of the data in DSNs, they are less effective to the problems where temporal dependencies exist in the input data. To embed the temporal information in DSNs, Recurrent Deep Stacking Networks (R-DSNs) \cite{Palangi2014} combined the advantages of DSNs and Recurrent neural networks (RNN). Unlike RNN which uses Back Propagation through time for training the network, R-DSNs use Echo State Network (ESN) to initialize the weights and then fine-tuning them via batch-mode gradient descent. A stacked extreme learning machine was proposed in \cite{Zhou2015}. Here, at each level of the network ELM with the reduced number of hidden nodes was used to solve the large scale problems. The number of hidden nodes was reduced via the principal component analysis (PCA) reduction technique.  Keeping in view the efficiency of stacked models, the number of stacked models based on support vector machine have been proposed \cite{Wang2019a, Wang2019b, Li2019}. Traditional models like Random Forests have also been extended to deep architecture, known as deep forests \cite{zhou2017deep}, via stacking concept.

In addition to DSNs, there are some novel network architectures proposed based on the stacking method, \citet{low2019stacking} contributed a stacking-based deep neural network (S-DNN) which is trained without a backpropagation algorithm. \citet{kang2020novel} presented a model by stacking conditionally restricted Boltzmann machine and deep neural network, which achieved significant superior performance with fewer parameters and fewer training samples.

\subsection{\textbf{Negative Correlation Based Deep Ensemble Methods}}
Negative correlation learning (NCL) \cite{Liu1999} is an important technique for training the learning algorithms. The main concept behind the NCL is to encourage diversity among the individual models of the ensemble to learn the diverse aspects of the training data. NCL minimizes the empirical risk function of the ensemble model via minimization of error functions of the individual networks. NCL \cite{Liu1999} was evaluated for regression as well as classification tasks. The evaluation used different measures like simple averaging and winner-takes-all measures on classification tasks and simple average combination methods for regression problems. The authors figured out that winner-takes-all is better as compared to simple averaging in NCL ensemble models.

\citet{shi2018crowd} proposed deep negative correlation learning architecture for crowd counting known as D-ConvNet i.e. decorrelated convolutional networks. Here, counting is done based on regression-based ensemble learning from a pool of convolutional feature mapped weak regressors. The main idea behind this is to introduce the NCL concept in deep architectures. Robust regression via deep NCL \cite{Zhang2019b} is an extension of \cite{shi2018crowd} in which theoretical insights about the Rademacher complexity are given and extended to more regression-based problems.

\citet{buschjager2020generalized} formulated a generalized bias-variance decomposition method to control the diversity and smoothly interpolates. They present the Generalized Negative Correlation Learning (GNCL) algorithm, which can encapsulate many existing works in literature and achieve superior performance.

The NCL can also be employed for incremental learning tasks. \citet{10.1007/978-3-540-72523-7_49} employed a dynamically modified weighted majority voting strategy to combine the sub-classifiers. \citet{TANG20092796} proposed a negative correlation learning (NCL) based approach for ensemble incremental learning. 

\subsection{\textbf{Explicit / Implicit Ensembles}}
Ensembling of deep neural networks doesn't seem to be an easy option as it may lead to increase in computational cost heavily due to the training of multiple neural networks. High performance hardware's with GPU acceleration may take weeks of weeks to train the deep networks. Implicit/Explicit ensembles obtain the contradictory goal wherein a single model is trained in such a manner that it behaves like ensemble  of training multiple neural networks without incurring additional cost or to keep the additional cost as minimum as possible. 
 Here, the training time of an ensemble is same as the training time of a single model. In implicit ensembles, the model parameters are shared and the single unthinned network at test times approximates the model averaging of the ensemble models.  However, in explicit ensembles model parameters are not shared and the ensemble output is taken as the combination of the predictions of the ensemble models via different approaches like majority voting, averaging and so on.  

Dropout \cite{srivastava2014dropout} creates an ensemble network by randomly dropping out  hidden nodes from the network during the training of the network. 
During the time of testing, all nodes are active. 
Dropout provides regularization of the network to avoid overfitting and introduces sparsity in the output vectors. Overfitting is reduced as it trains exponential number of models with shared weights and provides an implicit ensemble of networks during testing. Dropping the units randomly  avoids  coadaptation of the units by making the presence of a particular unit unreliable. The network with dropout takes $2-3$ times more time for training as compared to a standard neural network. Hence, a  balance is to be set appropriately between the training time of the network and the overfitting.   
Generalization  of DropOut was given in DropConnect \cite{pmlr-v28-wan13}. Unlike DropOut which drops each output unit, DropConnect randomly drops each connection and hence, introduces sparsity in the weight parameters of the model.
Similar to DropOut, DropConnect creates an implicit ensemble during test time by dropping out the connections (setting weights to zero) during training. 
Both DropOut and DropConnect suffer from high training time. To alleviate this problem, deep networks with Stochastic depth \cite{huang2016deep} aimed to reduce the network depth during training while keeping it unchanged during testing of the network. Stochastic depth is an improvement on ResNet \cite{he2016deep} wherein residual blocks are randomly dropped during training and bypassing these transformation blocks connections via skip connections. 
Swapout \cite{Singh} is a generalization of DropOut and Stochastic depth.
Swapout involves dropping of individual units or to skip the blocks randomly.
Embarking on a distinctive approach of reducing the test time, distilling the knowledge in a network \cite{hinton2015distilling} transferred the ``knowledge" from ensembles to a single model.  Gradual DropIn or regularised DropIn \cite{smith2016gradual} of layers starts from a shallow network wherein the layers are added gradually. DropIN trains the exponential number of thinner networks, similar to DropOut, and also shallower networks. 

All the aforementioned methods provided an ensemble of networks by sharing the weights.
There have been attempts to explore explicit ensembles in which models do not share the weights. Snapshot ensembling \cite{huang2017snapshot} develops an explicit ensemble without sharing the weights. The authors exploited good and bad local minima and let the stochastic gradient descent (SGD) converge $M$-times to local minima along the optimization path and take the snapshots only when the model reaches the minimum. These snapshots are then ensembled by averaging at multiple local minima for object recognition. The training time of the ensemble is the same as that of the single model. The ensemble out is taken as the average of the output of the snapshot outputs at multiple local minimas.
Random vector functional link network \cite{pao1994learning,malik2022random} has also been explored for creating the explicit ensembles \cite{katuwal2019random} where different random initialization of the hidden layer weights in a hierarchy diversifies the ensemble predictions.

Explicit/implicit produce ensembles out of a single network at the expense of base model diversity \cite{cao2020ensemble} as the lower level features across the models are likely to be the same. To alleviate this issue, branching based deep models \cite{han2017branchout} branch the network to induce more diversity. Motivated by different initializations of the neural networks leads to different local minima,  \citet{xue2021deep} proposed  deep ensemble model wherein ensemble of fully convolution neural network  over multiloss module  with coarse fine compensation module resulted in better segmentation of central serous chorioretinopathy lesion. Multiple neural networks with different initializations, multiple loss functions resulted in better diversity in an ensemble.

\begin{table}[]
    \centering
    \begin{tabular}{|c|c|c|}
    \hline
   Year& Authors& Contribution\\
    \hline
    2013&\citet{pmlr-v28-wan13}&  Introduced DropConnect (Random skipping of connections)\\
       2014& \citet{srivastava2014dropout} & Introduced Dropout  (Random skipping of units)\\
         2016& \citet{huang2016deep} &Deep networks with Stochastic depth (Random skipping of blocks) \\
        2016& \citet{Singh}& Introduced Swapout (Hybrid of Dropout and Stochastic depth approach)\\
         \hline
    \end{tabular}
    \caption{Implicit / Explicit ensembles}
    \label{tab:Implicit_Explicit_ensembles}
\end{table}

\subsection{\textbf{Homogeneous \& Heterogeneous ensembles}}
\begin{figure}[htb!]
    \centering
    \includegraphics[width=.75\textwidth]{Figures/hoe_Updated.png}
    \caption{Homogeneous ensemble (HOE) has models based on the same algorithm, but each individual model are fed with distinct datasets.}
    \label{Fig:hoe}
    
    \includegraphics[width=.75\textwidth]{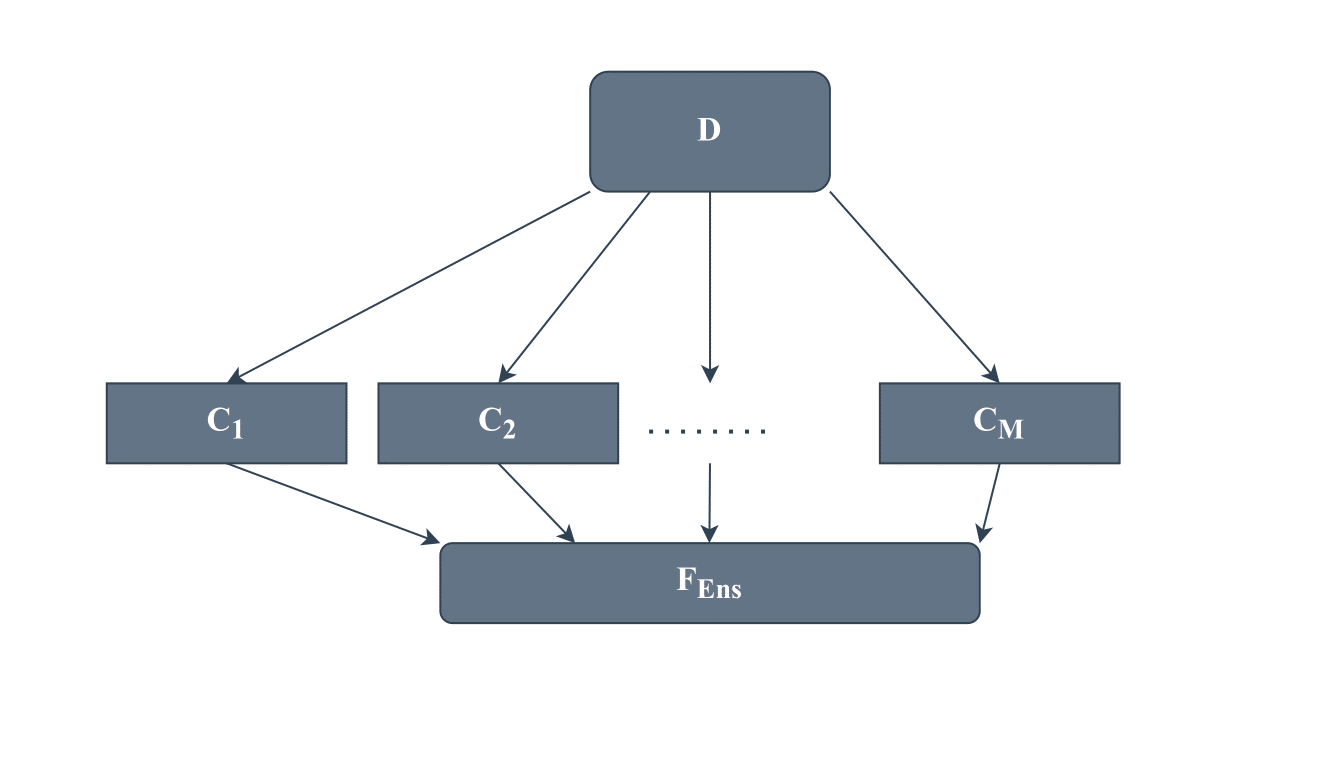}
    \caption{The components in heterogeneous ensemble (HEE) share the same dataset but consists of various algorithms.}
    \label{Fig:hee}
    
\end{figure}

Homogeneous ensemble (HOE) and heterogeneous ensemble (HEE) involve training a group of base learners either from the same family or different families, as shown in Fig.~\ref{Fig:hoe} and Fig.~\ref{Fig:hee}, respectively. Hence, each model of an ensemble must be as diverse as possible, and each base model must perform better than the random guess. The base learner can be a decision tree, neural network, or any other learning model. 

In homogeneous ensembles, the same base learner is used multiple times to generate the family of base classifiers. However, the key issue is to train each base model such that the ensemble model is as diverse as possible, i.e. no two models are making the same error on a particular data sample. The two most common ways of inducing randomness in a homogeneous ensemble are either sampling of the training set multiple times, thereby training each model on a different bootstrapped sample of the training data or sampling the feature space of the training data and train each model on a different feature subset of the training data. In some ensemble models like Random forest \cite{breiman2001random} used both these techniques for introducing diversity in the ensemble of decision trees. 
In neural networks, training models independently with different initialization of the models also induces diversity.
However, deep learning models have high training costs and hence, training of multiple deep learning models is not a feasible option. Some attempts, like horizontal vertical voting of deep ensembles \cite{xie2013horizontal} have been made to obtain ensembles of deep models without independent training.
Temporal ensemble \cite{laine2016temporal} trains multiple models with different input augmentation, different regularisation and different training epochs. Training of multiple deep neural networks for image classification \cite{ciregan2012multi}  and for disease prediction \cite{grassmann2018deep} showed that better performance is achieved via an ensemble of multiple networks and averaging the outputs.
Despite these models, training multiple deep learning models for ensemble is an uphill task as millions or billions of parameters need to be optimized. Hence, some studies have used deep learning in combination with traditional models to build heterogeneous ensemble models, enjoying the benefits of lower computation and higher diversity. Heterogeneous ensemble for default prediction \cite{li2018heterogeneous} is an ensemble of the extreme gradient boosting, deep neural network and logistic regression. Heterogeneous ensemble for text classification \cite{kilimci2018deep} is an ensemble of multivariate Bernoulli naïve Bayes (MVNB), multinomial naïve Bayes (MNB), support vector machine (SVM), random forest (RF), and convolutional neural network (CNN) learning algorithms. Using different perspectives of data, model and decision fusion, heterogeneous deep network fusion \cite{tabik2020mnist} showed that complex heterogeneous fusion architectures are more diverse and hence, show better generalization performance. Furthermore,~\citet{seijo2017ensemble} employed both homogeneous and heterogeneous ensembles for feature selection.
\citet{10.1007/978-3-642-17313-4_1} suggested that the heterogeneous bagging based ensemble strategy performs better than boosting based Learn++ algorithms and some other NCL methods. 
Other examples that employed homogeneous ensemble methods were used to deal with the presence of incremental tasks, such as concept drift \cite{minku2009impact}, power load forecasting \cite{QIU2018182, Grmanov2016IncrementalEL}, myoelectric prosthetic hands surface electromyogram characteristics \cite{7518678}, etc. \citet{DAS2016170} proposed an ensemble incremental learning with pseudo-outer-product fuzzy neural network for traffic flow prediction, real-life stock price, and volatility predictions, etc. 

\subsection{\textbf{Decision Fusion Strategies}}
Ensemble learning trains several base learners and aggregates the outputs of base learners using some rules. The rule used to combine the outputs determines the effective performance of an ensemble. Most of the ensemble models focus on the ensemble architecture followed by their naive averaging to predict the ensemble output. However, naive averaging of the models, followed in most of the ensemble models, is not data adaptive and leads to less optimal performance  \cite{ju2018relative} as it is sensitive to the performance of the biased learners. As there are billions of hyperparameters in deep learning architecture, the issue of overfitting may lead to the failure of some base learners. Hence, to overcome these issues, approaches like Bayes optimal classifier and super learner have been followed  \cite{ju2018relative}. 

The different approaches followed in the literature for combining the outputs of the ensemble models are:

\subsubsection{\textbf{Unweighted Model Averaging}}
Unweighted averaging  of the outputs of the base learners in an ensemble is the most followed approach for fusing the decisions in the literature. Here, the outcomes of the base learners are averaged to get the final prediction of the ensemble model. Deep learning architectures have high variance and low bias, thus, simple averaging of the ensemble models improve the generalization performance due to the reduction of the variance among the models.

The averaging of the base learners is performed either on the outputs of the base learners directly or on the predicted probabilities of the classes via softmax function: 
\begin{align}
    P_i^j=softmax^j(O_i)=\frac{O_i^j}{\sum_{k=1}^K exp(O_k^j)}
\end{align}
where $P_i^j$ is the probability outcome of the $i^{th}$ unit on the $j^{th}$ base learner, $O_i^j$ is the output of the $i^{th}$ unite of the  $j^{th}$  base learner and $K$ is the number of the classes.

Unweighted averaging is a reasonable choice when the performance of the base learners is comparable, as suggested in \cite{he2016deep,simonyan2014very,szegedy2015going}. However, when the ensemble contains heterogeneous base learners naive unweighted averaging may result in suboptimal performance as it is affected by the performance of the weak learners and the overconfident learners \cite{ju2018relative}. The adaptive metalearner should be good enough to adaptively combine the strengths of the base learners as some learners may have lower overall performance but maybe good at the classification of certain subclasses and hence, leading to better overall performance.

\subsubsection{\textbf{Majority Voting}}
Similar to unweighted averaging, majority voting combines the outputs of the base learners. However, instead of taking the average of the probability outcomes, majority voting counts the votes of the base learners and predicts the final labels as the label with the majority of votes. In comparison to unweighted averaging, majority voting is less biased towards the outcome of a particular base learner as the effect is mitigated by majority vote count. However, favouring of a particular event by most of the similar base learners or dependent base learners leads to the dominance of the event in the ensemble model. In majority voting, the analysis by \citet{kuncheva2003limits} showed that the pairwise dependence among the base learners plays an important role and for the classification of images, the prediction of shallow networks   is more diverse as compared to the deeper networks \cite{choromanska2015loss}. Hence,  \citet{ju2018relative} hypothesised that the performance of the majority voting based shallows ensemble models is better as compared to the majority based deep ensemble models.

Voting methods have also started to be integrated with semi-supervised deep learning. \citet{li2017semi} proposed an ensemble semi-supervised deep acoustic models for in automatic speech recognition. \citet{wang2019enaet} explored an ensemble self-learning method to enhance semi-supervised performance and extract adverse drug events from social media in \cite{liu2018ssel}. In the semi-supervised classification area, the author proposed a deep coupled ensemble learning method which is combined with complementary consistency regularization and gets the state of the art performance in \cite{li2019semi}. Some results have also been achieved with semi-supervised ensemble learning on some datasets where the annotation is costly. \citet{pio2014integrating} employed an ensemble method to improve the reliability of miRNA:miRNA predicted interactions.

Furthermore, the multi-label classification \cite{tsoumakas2007multi} problem is also a major point addressed by the voting method, a typical application is the RAndom $k$-labELsets (RAKEL) algorithm \cite{tsoumakas2007random}. The author trained several single-label classifiers using small random subsets of actual labels. Then the final output is carried out by a voting scheme based on the predictions of these single classifiers. There are also many variants of RAKEL proposed in recent years \cite{moyano2019evolutionary, kimura2016fast,wang2021active}. \citet{10.1007/978-3-642-23808-6_15} proposed a solution for multi-label ensemble learning problem, which construct several accurate and diverse multi-label based basic classifiers and employ two objective functions to evaluate the accuracy and diversity of multi-label base learners. Another work \cite{LI2013269} proposed an ensemble multi-label classification framework based on variable pairwise constraint projection. \citet{xia2020multi} proposed a weighted stacked ensemble scheme that employs the sparsity regularization to facilitate classifier selection and ensemble construction. Besides, there are many applications of ensemble multi-label methods. Some publications employ multi-label ensemble classifiers to explore the protein, such as protein subcellular localization \cite{guo2016human}, protein function prediction \cite{yu2012transductive}, etc. The Muli-label classifier is also utilized in predicting the drug side effects \cite{zhang2015predicting}, predicting the gene prediction\cite{schietgat2010predicting}, etc. Moreover, there is another critical ensemble multi-label algorithm called ensemble classifier chains (ECC) \cite{read2011classifier}. This method involves binary classifiers linked along a chain. The first classifier is trained using only the input data, and then each subsequent classifier is trained on the input space and all previous classifiers in the chain. The final prediction is obtained by the integration of the predictions and selection above a manually set threshold. \citet{chen2017ensemble} propose an ensemble application of convolutional and recurrent neural networks to capture both the global and local textual semantics and to model high-order label correlations.

\subsubsection{\textbf{Bayes Optimal Classifier}}
In Bayesian method, hypothesis $h_j$ of each base learner with the conditional distribution of target label $t$ given $x$. Let $h_j$ be the hypothesis generated on the training data $D$ evaluated on test data $(x,t)$, mathematically,  $h_j(t|x)=P[y|x,h_j,D]$. With Bayes rule, we have
\begin{align}
    P(t|x,D)	\propto \sum_{h_j}P[t|h_j,x,D] P[D|h_j] P[h_j]
\end{align}
and the Bayesian Optimal classifier is given as:
\begin{align}
    \underset{t}{argmax}~~ \sum_{h_j}P[t|h_j,x,D] P[D|h_j] P[h_j],
\end{align}
where $P[D|h_j]=\Pi_{(t,x)\in D} h_j(t|x)$ is the likelihood of the data under $h_j$. However, due to overfitting issues this might be not a good measure. Hence, training data is divided into two sets-one for training the model and the other for evaluating the model. Usually validation set is used to tune the hyperparameters of the  model.

Choosing prior probabilities in Bayes optimal classifier is difficult and hence, usually set to uniform distribution for simplicity. With a large sample size, one hypothesis tends to give larger posterior probabilities than others and hence the weight vector is dominated by a single base learner and hence Bayes optimal classifier would behave as the discrete superlearner with a negative likelihood loss function.  

\subsubsection{\textbf{Stacked Generalization}}
Stacked generalization \cite{wolpert1992stacked} works by deducing the biases of the generalizer(s) with respect to a provided learning set. To obtain the good linear combination of the base learners in regression, cross-validation data and least squares under non-negativity constraints was used to get the optimal weights of combination \cite{breiman1996stacked}. Consider the linear combination of the predictions of the  base learners $f_1,f_2,\cdots,f_m$ given as:
\begin{align}
    f_{stacking}(x)=\sum_{j=1}^m w_jf_j(x)
\end{align}
where $w$ is the optimal weight vector learned by the meta learner.

\subsubsection{\textbf{Super Learner}}
Inspired by the cross validation for choosing the optimal classifier, \citet{van2007super} proposed super learner which is weighted combination of the predictions of the base learner. Unlike the stacking approach, it uses cross validation approach to select the optimal weights for combining the predictions of the base learners. 

With smaller datasets, cross validation approach can be used to optimize the weights. However, with the increase in the size of the data and the number of base learners in the model, it may not be a feasible option. Instead of optimizing the V-fold cross validation, single split cross validation can also be used for optimizing the weights for optimal combination \cite{ju2019propensity}. In deep learning models, usually, a validation set is used to evaluate the performance instead of using the cross validation.

Another application field for super learner is in Reinforcement Learning 
With the development of Deep learning, some researchers have implemented deep reinforcement learning, which combines deep learning with a Q-learning algorithm \cite{mnih2013playing}. Ensemble methods in deep Q learning have decent performance. \citet{chen2018ensemble} proposed an ensemble network architecture for deep reinforcement learning. The integrated network includes Temporal Ensemble and Target Values Ensemble. Develop a human-like chat robot is a challenging job, by incorporating deep reinforcement learning and ensemble method,  \citet{cuayahuitl2019ensemble}  integrated 100 deep reinforcement learning agents, the agents are trained based on clustered dialogues. They also demonstrate the ensemble of DRL agents has better performance than the single variant or Seq2Seq model. Stock trading is another topic where ensemble deep reinforcement learning has achieved a promising result. \citet{carta2020multi} found the single supervised classifier is inadequate to deal with the complex and volatile stock market. They employed hundreds of neural networks to pre-process the data, then they combined several reward-based meta learners as a trading agency. Moreover, \citet{yang2020deep} trained an ensemble trading agency based on three different metrics: Proximal Policy Optimization (PPO), Advantage Actor-Critic (A2C), and Deep Deterministic Policy Gradient (DDPG). The ensemble strategy combines the advantages of the three different algorithms. Besides, some researchers try to use ensemble strategy to solve the disease-prediction problem. The proposed model in \cite{tang2016inquire} consists of several sub-models which are in response to different anatomical parts.


\subsubsection{\textbf{Consensus}}

\begin{figure}[htb!]
    \centering
    \includegraphics[width=.8\textwidth]{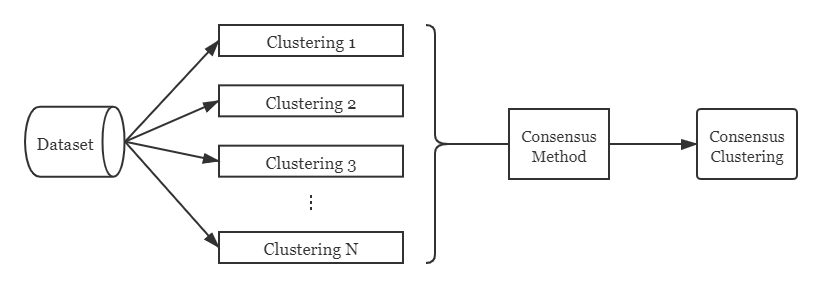}
    \caption{The process of consensus clustering. An ensemble of different clustering results can be combined by a consensus approach.}
    \label{Fig:cc}
    
\end{figure}

Unsupervised learning is another group of machine learning techniques. The fundamental difference between it and supervised learning is that unsupervised learning usually handles training samples without corresponding labels. Therefore, the primary usage of unsupervised learning is to do clustering. The reason why ensemble methods are employed is to combine some weak clusters into strong one. To create diverse clusters, several approaches can be applied: using different sampling data, using different subsets of the original features, and employing different clustering methods \cite{senbabaouglu2014critical}. Sometimes, even some random noise can be added to these base models to increase randomness, which is good for ensemble methods according to \cite{bian2007diversity}. After receiving all the outputs from each cluster, various consensus functions can be chosen to obtain the final output based on the user's requirement \cite{vega2011survey}. The ensemble clustering is also known as consensus clustering Fig.~\ref{Fig:cc}.

\citet{zhou2006clusterer} explored ensemble methods for unsupervised learning and developed four different approaches to combine the outputs of these clusters. In recent years, some new ensemble clustering methods have been proposed that illustrated the priority of ensemble learning \cite{huang2017locally,zheng2010hierarchical,huang2016ensemble}. Most of the clustering ensemble method is based on the co-association matrix solution, which can be regarded as a graph partition problem. Besides, there is some research focus on integrating the deep structure and ensemble clustering method. \citet{liu2015spectral, liu2016infinite} firstly showed that ensemble unsupervised representation learning with deep structure can be applied in large scale data. Then the author combined the method with auto-encoder and extends it to the vision field. \citet{shaham2016deep} first demonstrated that some crowdsourcing algorithms can be replaced by a Restricted Boltzmann Machine with a single hidden neuron, then propose an RBM-based Deep Neural Net (DNN) used for unsupervised ensemble learning.  The unsupervised ensemble method also makes some contribution to the field of Natural Language Processing. \citet{alami2019enhancing} demonstrated that the ensemble of unsupervised deep neural network models that use Sentence2Vec representation as the input has the best performance according to the experiments. \citet{hassan2019uests} proposed a module that includes four semantic similarity measures, which improves the performance on the semantic textual similarity (STS) task. The unsupervised ensemble method is also widely used for tasks that lack annotation, such as the medical image. \citet{ahn2019unsupervised} proposed an unsupervised feature learning method integrated ensemble approach with a traditional convolutional neural network. \citet{lahiri2016deep} employed unsupervised hierarchical feature learning with ensemble sparsely autoencoder on retinal blood vessels segmentation task, meanwhile, \citet{liu2019unsupervised} also propose an unsupervised ensemble architecture to automatically segment retinal vessel. Besides, there are also some ensemble deep methods working on localization predicting for long non-coding RNAs \cite{cao2018lnclocator}. \citet{hu2022representation} extended the ensemble random vector functional link to unsupervised tasks. The authors employ manifold regularization to re-represent the original features, and then use the Kuhn-Munkre algorithm with consensus clustering to ensemble the clustering results from multiple hidden layers.

\subsubsection{\textbf{Query-By-Committee}}

\begin{figure}[htbp]
    \centering
    \includegraphics[width=.8\textwidth]{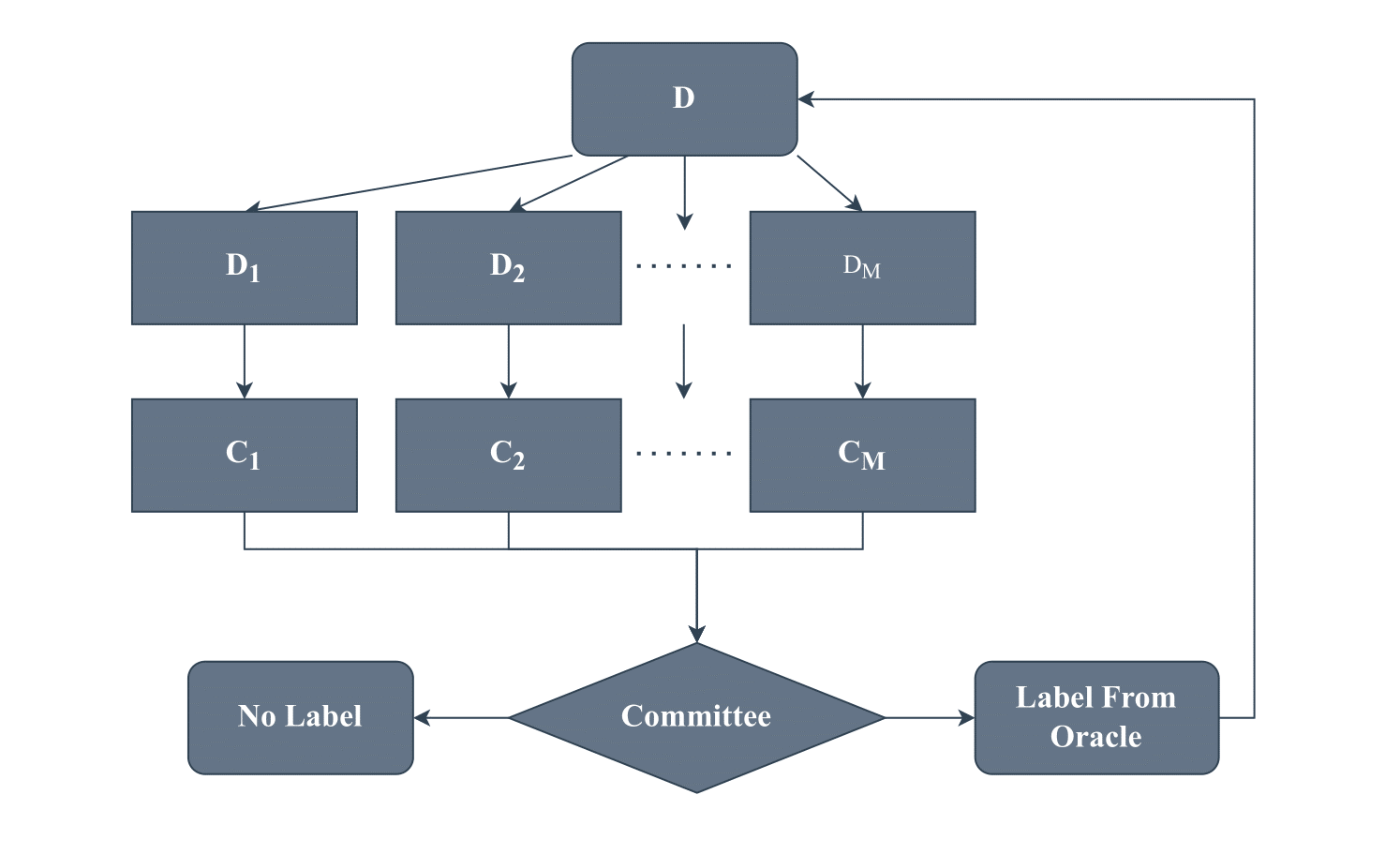}
    \caption{Query-by-committee in Active Learning. Sampling with replacement is used to partition the labeled training data set into training splits. The committee determines whether to label the data based on the output of several algorithms.}
    \label{Fig:qbc}
    
\end{figure}

Active Learning is another popular topic in the deep learning area, which is also often used in conjunction with semi-supervised learning and ensemble learning. The key sight of this is to make the algorithm learning from less annotated data. Some conventional active learning algorithms, such as Query-By-Committee (as shown in Fig~\ref{Fig:qbc}), have already adopted the idea of ensemble learning.  \citet{melville2003constructing,melville2004diverse} explored an ensemble method that builds a diverse committee. \citet{beluch2018power} discussed the power of ensembles for active learning is significantly better than Monte-Carlo Dropout and geometric approaches. \citet{sharma2018dti} show some applications in drug-target interaction prediction. Ensemble active learning is also available to conquer the concept drift and class imbalance problem \cite{8543581}.

\section{Applications}
\label{sec:Applications}
\begin{figure}
    \centering
    \includegraphics[width=0.9\textwidth]{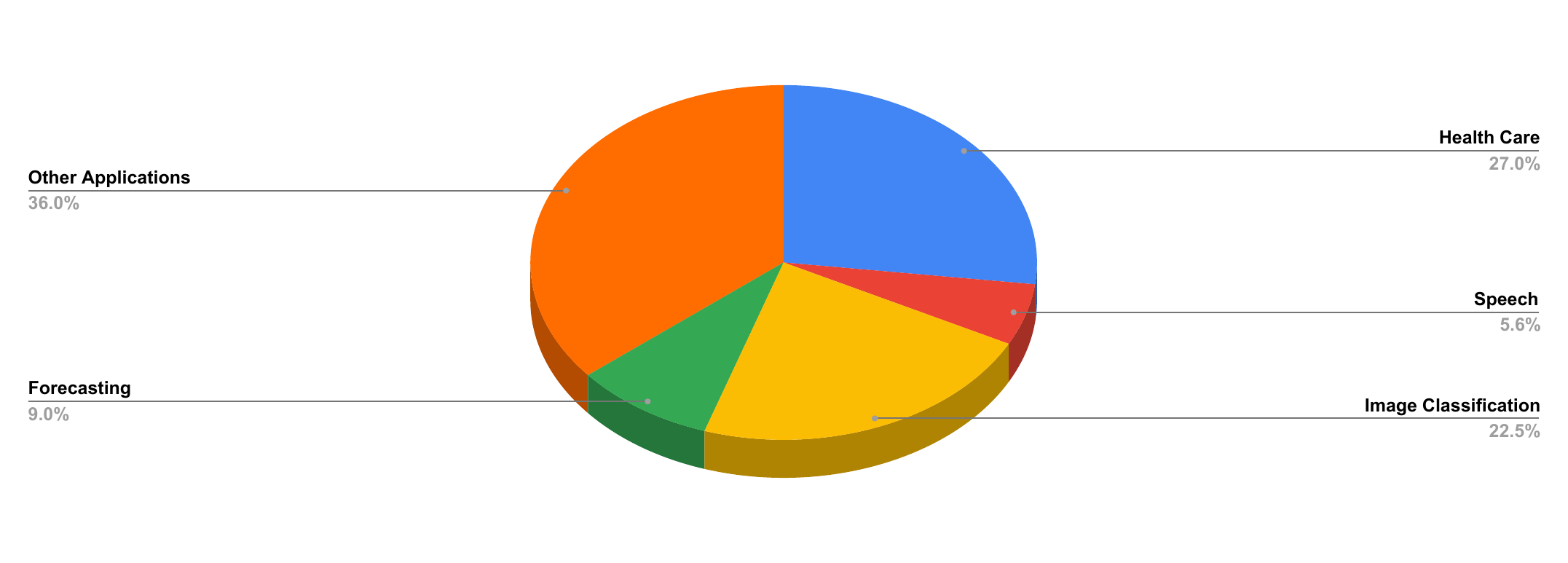}
    \caption{Ensemble-based approach in different areas. Data from Tables~\ref{tab:Health Care} to~\ref{tab:Other applications}.}
    \label{fig:Applications}
\end{figure}

\begin{figure}[bthp!]
    \centering
    \includegraphics[width=0.9\textwidth]{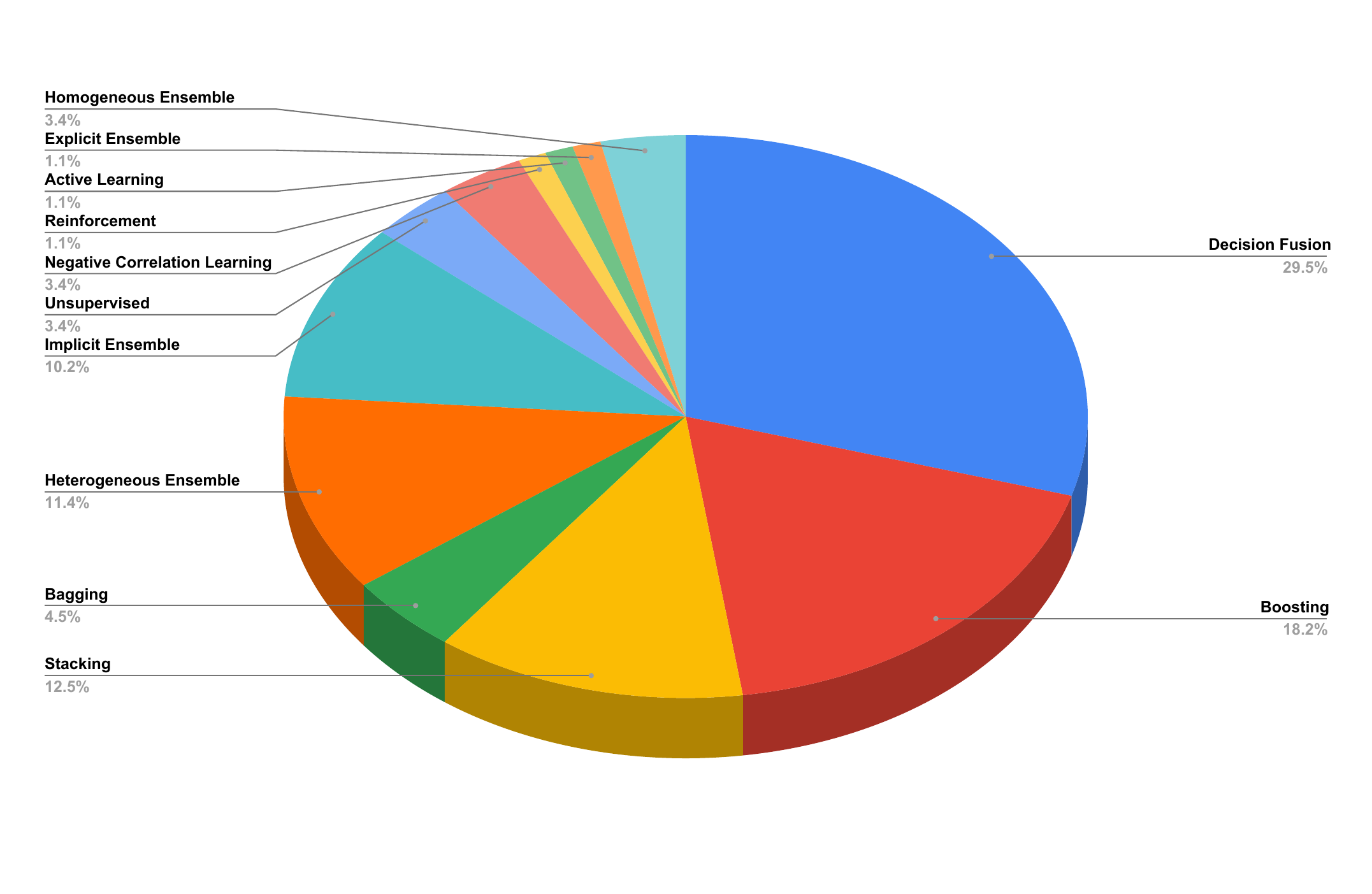}
    \caption{Analysis of the applications of various ensemble methods. Data from Tables~\ref{tab:Health Care} to~\ref{tab:Other applications}.}
    \label{fig:Ensemble Strategy}
\end{figure}

In this section, we briefly present the applications of deep ensemble models across different domains in a tabular form.
 Ensemble deep models have been implemented in several domains and therefore, in broad sense, we have classified the application domains into five categories, i.e., health care, speech, image classification, forecasting and the rest models are listed in others category. Table \ref{tab:Health Care} gives the information about the ensemble deep models that have been implemented in health care domain. Here, several papers are based on heterogeneous ensemble technique. It reveals that using different family's models into a single frame perform better in health care domain. 
 Recently, ensemble deep techniques have been successful and have shown good performance in health care domain. The models which have been implemented for speech task have been given in Table \ref{tab:Speech} and most of the ensemble approaches are based on stacking technique. Table \ref{tab:Image  Classification} contains the ensemble deep models that have been implemented in speech areas. Models that have been implemented in forecasting and other domains have been given in Table \ref{tab:Forecasting} and Table \ref{tab:Other applications}, respectively.
Fig. \ref{fig:Applications} shows the percentages of the application domains. 
The statistics reveals that different ensemble deep techniques have been used in different areas. A larger number of models, i.e. $27\%$ of the ensemble deep models, have been implemented in health care domain and $5.6\%$ percent of the models for speech application and $22.5\%$ of the models for image classification task. Moreover,  $9\%$ models have been used in forecasting and $36\%$ in other applications areas, i.e., information retrieval, emotion recognition, text categorization and so on.
Fig. \ref{fig:Ensemble Strategy} shows the ensemble strategies in percentage. 
In ensemble learning, there are several ways to integrate the outcomes of the models in an ensemble. In the literature, researchers have proposed different techniques of decision fusion according to different areas of application. Bagging, boosting and stacking are the classical ensemble techniques. Based on these three techniques, researchers have developed several other techniques also. Boosting ($18.2\%$), stacking (12.5\%) and bagging (4.5\%) techniques have been implemented in ensemble deep framework. Heterogeneous and implicit ensemble  are also popular for making an efficient ensemble  model and their contribution are as follows $11.4\%$ and $10.2\%$, respectively. The rest ensemble techniques are, i.e. unsupervised ($3.4\%$), NCL (3.4\%), reinforcement (1.1\%),
active learning ($1.1\%$), explicit ensemble (1.1\%) and homogeneous ensemble (3.4\%).

\begin{table}[]
    \centering
    \resizebox{\textwidth}{!}{
    \begin{tabular}{|l|l|p{0.7\linewidth}|l|l|}
    \hline
    Year & Author&Title&Approach&Area\\
    \hline

2014 &\citet{zheng2014hibag}& HIBAG—HLA genotype imputation with attribute bagging  & Bagging & Genotype Imputation \\ 
2014&\citet{Cortes2014}&Deep Boosting & Boosting&Classification \\
2015 &\citet{zhang2015predicting}& Predicting drug side effects by multi-label learning and ensemble learning  & Decision Fusion & Predict the drug side effects \\ 
2016 &\citet{guo2016human}& Human protein subcellular localization with integrated source and multi-label ensemble classifier. & Decision Fusion &  Protein subcellular localization prediction\\ 
2016 &\citet{lahiri2016deep}& Deep neural ensemble for retinal vessel segmentation in fundus images towards achieving label-free angiography  & Decision Fusion & Medical image segmentation\\ 
2017 &\citet{cabria2017mri}& MRI segmentation fusion for brain tumor detection & Heterogeneous ensemble & MRI segmentation\\
2018&\citet{grassmann2018deep}& A deep learning algorithm for prediction of age-related eye disease study severity scale for age-related macular degeneration from color fundus photography  & Homogeneous ensemble & Disease prediction\\ 
2018 &\citet{cao2018lnclocator}& The lnclocator: a subcellular localization predictor for long non-coding RNAs based on a stacked ensemble classifier  & Decision Fusion & Subcellular localization predictor \\
2018 & \citet{sharma2018dti}&Be-dti’: Ensemble framework for drug target interaction prediction using dimensionality reduction and active learning & Active learning & Drug target interaction prediction \\ 
2019 & \citet{ahn2019unsupervised} &Unsupervised feature learning with k-means and an ensemble of deep convolutional neural networks for medical image classification & Decision Fusion &  Medical image classification \\
2019 &\citet{liu2019unsupervised}& Unsupervised ensemble strategy for retinal vessel segmentation.  & Unsupervised & Medical image classification\\
2020&\citet{shalbaf2021automated}& Automated detection of COVID-19 using ensemble of transfer learning with deep convolutional neural network based on CT scans & Heterogeneous Ensemble &Detection of COVID-19\\
2020& \citet{ali2020smart}& A smart healthcare monitoring system for heart disease prediction based on ensemble deep learning and feature fusion&Boosting &Heart disease prediction\\
2021& \citet{zhou2021ensemble}& The ensemble deep learning model for novel COVID-19 on CT images& Heterogeneous Ensemble& Detection of COVID-19\\
2021&\citet{li2020intelligent}&Intelligent Fault Diagnosis by Fusing Domain Adversarial Training and Maximum Mean Discrepancy via Ensemble Learning &Heterogeneous Ensemble&Fault diagnosis\\
2021&\citet{das2021automatic}&Automatic COVID-19 detection from X-ray images using ensemble learning with convolutional neural network
&Heterogeneous Ensemble&Detection of COVID-19\\
2022&\citet{sukegawa2022identification}&Identification of osteoporosis using ensemble deep learning model with panoramic radiographs and clinical covariates &Decision Fusion& Identification of osteoporosis\\
2022&\citet{gao2022vessel}& Vessel segmentation for X-ray coronary angiography using ensemble methods with deep learning and filter-based features &Boosting &Vessel segmentation\\
2022&\citet{rath2022improved}& Improved heart disease detection from {ECG} signal using deep learning based ensemble model &Heterogeneous Ensemble&Heart disease detection\\
2022& \citet{tanveer2021classification}&Classification of Alzheimer’s Disease Using Ensemble of Deep Neural Networks Trained Through Transfer Learning   &Heterogeneous Ensemble    &Classification of Alzheimer’s Disease\\
2022&\citet{rai2022hybrid}&Hybrid CNN-LSTM deep learning model and ensemble technique
for automatic detection of myocardial infarction using big ECG data &Heterogeneous Ensemble&Detection of myocardial infarction\\
2022&\citet{ganaie2022ensemble}&Ensemble deep random vector functional link network using privileged information for Alzheimer's disease diagnosis&Implicit ensemble& Diagnosis of Alzheimer's disease\\ \hline
    \end{tabular}}
    \caption{Applications in health care}
    \label{tab:Health Care}
\end{table}

\begin{table}[]
    \centering
    \resizebox{\textwidth}{!}{
    \begin{tabular}{|l|l|p{0.7\linewidth}|l|l|}
    \hline
    Year & Author&Title&Approach&Area\\
    \hline
         2012&\citet{tur2012towards}& Towards deeper understanding: Deep convex networks for semantic utterance classification  & Stacking & Semantic Utterance Classification \\
2012 &\citet{deng2012use}& Use of kernel deep convex networks and end-to-end learning for spoken language understanding  & Stacking & Spoken Language Understanding \\
2014 &\citet{deng2014ensemble}& Ensemble deep learning for speech recognition  & Stacking & Speech Recognition \\ 
2014 &\citet{Palangi2014}& Recurrent Deep-Stacking Networks for sequence classification  & Stacking & Sequence classification\\ 
2017&\citet{li2017semi}& Semi-supervised ensemble DNN acoustic model training  & Decision Fusion & Speech Recognition \\ 
\hline
    \end{tabular}}
    \caption{Applications in speech}
    \label{tab:Speech}
\end{table}

\begin{table}[]
    \centering
    \resizebox{\textwidth}{!}{
    \begin{tabular}{|l|l|p{0.7\linewidth}|l|l|}
    \hline
    Year & Author&Title&Approach&Area\\
    \hline
2012 &\citet{ciregan2012multi}& Multi-column deep neural networks for image classification & Homogeneous ensemble & Classification\\
2013 &\citet{pmlr-v28-wan13}& Regularization of Neural Networks using DropConnect  & Implicit ensemble & Image recognition\\ 
2014&\citet{srivastava2014dropout}& Dropout: a simple way to prevent neural networks from overfitting  & Implicit Ensemble & Computer vision, speech recognition \\ 
&&& & document classification and computational biology \\ 
2014 &\citet{liu2014facial}& Facial expression recognition via a boosted deep belief network& Boosting& Facial expression recognition\\
2015 &\citet{li2015sparse}& Sparse deep stacking network for image classification  & Stacking & Image Classification \\ 
2015& \citet{yang2015convolutional}&Convolutional channel features  & Boosting & Pedestrian detection, face detection,\\ 
&&& & edge detection and object proposal generation\\ 
2016&\citet{moghimi2016boosted}& Boosted Convolutional Neural Networks & Boosting & Classification\\ 
2016 &\citet{huang2016deep}& Deep networks with stochastic depth  & Implicit ensemble& Classification\\ 
2016 &\citet{he2016deep}& Deep residual learning for image recognition & Implicit ensemble &  classification, and  object detection\\ 
2016 &  \citet{Singh}&Swapout: Learning an ensemble of deep architectures& Implicit ensemble & classification\\ 
2016 &\citet{smith2016gradual}& Gradual dropin of layers to train very deep neural networks  & Implicit ensemble & Classification\\
2016 & \citet{laine2016temporal}& Temporal ensembling for semi-supervised learning & Homogeneous ensemble & Classification\\
2016 &\citet{tang2016inquire}& Inquire and diagnose: Neural symptom checking ensemble using deep reinforcement learning  & Decision Fusion & Inquire symptoms and diagnose diseases\\
2017 &\citet{huang2017snapshot}& Snapshot ensembles: train 1, get M for free & Explicit ensemble & Classification \\ 
2017&\citet{mosca2017deep}& Deep incremental boosting  & Boosting & Classification\\ 
2018 & \citet{beluch2018power}& The power of ensembles for active learning in image classification  & Decision Fusion & Image classification\\
2019 & \citet{amin2019ensemble}&Ensemble of CNN for multi-focus image fusion  & Decision Fusion & Image Classification\\
2019 &\citet{li2019semi}& Semi-supervised deep coupled ensemble learning with classification landmark exploration.  & Decision Fusion & Image classification \\
2020 &\citet{wang2020particle}& Particle swarm optimisation for evolving deep neural networks for image classification by evolving and stacking transferable blocks.  & Stacking & Image Classification\\
\hline
    \end{tabular}}
    \caption{Applications in image  classification}
    \label{tab:Image  Classification}
\end{table}

\begin{table}[]
    \centering
   \resizebox{\textwidth}{!}{
    \begin{tabular}{|l|l|p{0.7\linewidth}|l|l|}
    \hline
    Year & Author&Title&Approach&Area\\
    \hline
         2014 &\citet{qiu2014ensemble}& Ensemble deep learning for regression and time series forecasting  & Decision Fusion & Regression and Time Series Forecasting \\ 
2016 &\citet{Grmanov2016IncrementalEL}& Incremental ensemble learning for electricity load forecasting  & Decision Fusion & Electricity load forecasting\\ 
2017 & \citet{qiu2017empirical}&Empirical Mode Decomposition based ensemble deep learning for load demand time series forecasting  &  Decision Fusion & Load demand forecasting \\ 
2017 & \citet{liu2017flood}&A Flood Forecasting Model Based on Deep Learning Algorithm via Integrating Stacked Autoencoders with BP Neural Network  & Stacking & Flood Forecasting \\
2018 &\citet{QIU2018182}& Ensemble incremental learning random vector functional link network for short-term electric load forecasting.  & Decision Fusion & Electric load forecasting\\ 
2020 & \citet{carta2020multi}&A multi-layer and multi-ensemble stock trader using deep learning and deep reinforcement learning  & Implicit ensemble & Stock trader\\
2020 &\citet{yang2020deep}& Deep reinforcement learning for automated stock trading: An ensemble strategy  &  Decision Fusion  & Stock trading agency \\
2021 & \citet{bhusal2021deep} & Deep ensemble learning-based approach to real-time power system state estimation & Stacking & Electric Power\\
2022&\citet{singla2021ensemble}&An ensemble method to forecast 24-h ahead solar irradiance using
wavelet decomposition and BiLSTM deep learning network &Decision Fusion&Forecasting\\
\hline
    \end{tabular}}
    \caption{Applications in forecasting}
    \label{tab:Forecasting}
\end{table}

\begin{table}[]
    \centering
    \resizebox{\textwidth}{!}{
    \begin{tabular}{|l|l|p{0.7\linewidth}|l|l|}
    \hline
    Year & Author&Title&Approach&Area\\
    \hline
        2013 & \citet{deng2013deep}&Deep stacking networks for information retrieval  & Stacking & Information Retrieval \\ 
2014&\citet{Kuznetsov2014}& Multi-class deep boosting & Boosting&Classification\\ 
2014 &\citet{wang2014sentiment}& Sentiment classification The contribution of ensemble learning  & Bagging, Boosting & Sentiment classification \\ 
2015 &\citet{zareapoor2015application}& Application of Credit Card Fraud Detection: Based on Bagging Ensemble Classifier  & Bagging & Credit Card Fraud Detection \\ 
2016 &\citet{yin2017recognition} & Recognition of emotions using multimodal physiological signals and an ensemble deep learning model & Decision Fusion & Emotions Recognition \\  
2016 &\citet{liu2016infinite}&  A deep learning approach to unsupervised ensemble learning  & Decision Fusion & Clustering\\
2016&\citet{walach2016learning}& Learning to count with {CNN} boosting & Boosting & Object counting in images\\ 
2017& \citet{han2017incremental}&Incremental boosting convolutional neural network for facial action unit recognition  & Boosting& Facial action unit recognition\\ 
2017&\citet{chen2017ensemble}& Ensemble application of convolutional and recurrent neural networks for multi-label text categorization  & Decision Fusion & Text Categorization.\\
2017&\citet{opitz2017bier}& Bier-boosting independent embeddings robustly  & Boosting & Image retrieval\\ 
2018 & \citet{shi2018crowd}&Crowd Counting with Deep Negative Correlation Learning   & Negative correlation learning & Crowd Counting\\ 
2018 &\citet{kazemi2018novel} & Novel genetic-based negative correlation learning for estimating soil temperature & Negative correlation learning & Soil Temperature Estimation \\ 
2018 &\citet{randhawa2018credit}& Credit Card Fraud Detection Using AdaBoost and Majority Voting  & Boosting & Credit Card Fraud Detection \\ 
2018 &\citet{sun2018sparse}& Sparse Deep Stacking Network for Fault Diagnosis of Motor  & Stacking & Fault Diagnosis \\ 
2018&\citet{chen2018deep}& Deep boosting for image denoising & Boosting & Image denoising\\ 
2018 &\citet{li2018heterogeneous}&Heterogeneous ensemble for default prediction of peer-to-peer lending in China & Heterogeneous ensemble& Default prediction\\
2018 &\citet{kilimci2018deep}& Deep learning-and word embedding-based heterogeneous classifier ensembles for text classification & Heterogeneous ensemble & Classification\\
2018 &\citet{liu2018ssel}& SSEL-ADE:  a semi-supervised  ensemble  learning  framework  for extracting adverse drug events from social media  & Decision Fusion & Extracting adverse drug events \\
2019 &\citet{martin2019android}& Android malware detection through hybrid features fusion and ensemble classifiers: The AndroPyTool framework and the OmniDroid dataset  & Decision Fusion & Android malware detection\\
2019 &\citet{cuayahuitl2019ensemble} & Ensemble-based deep reinforcement learning for chatbots & Reinforcement & Chat robot \\ 
2019& \citet{chen2019real} & Real-world image denoising with deep boosting& Boosting & Image denoising\\ 
2019&\citet{waltner2019hibster}& HiBsteR: Hierarchical Boosted Deep Metric Learning for Image Retrieval  & Boosting & Image Retrieval \\ 
2019 &\citet{wang2019adaboost}& Adaboost-based security level classification of mobile intelligent terminals  &  Adaboost & Security Level Classification\\
2019 &\citet{chen2019novel}& Novel Hybrid Integration Approach of Bagging-Based Fisher’s Linear Discriminant Function for Groundwater Potential Analysis  & Bagging & Groundwater Potential Analysis\\
2019 & \citet{katuwal2019random}&Random vector functional link neural network based ensemble deep learning  & Implicit ensemble & Classification\\
2019 & \citet{zhang2019deep}& Deep stacked hierarchical multi-patch network for image deblurring & Stacking & Deblurring Image\\
2019 &\citet{alami2019enhancing}&  Enhancing unsupervised neural networks based text summarization  with word embedding and ensemble learning  & Decision Fusion  & Text summarization \\
2019 & \citet{hassan2019uests}& Uests:  An unsupervised ensemble semantic textual similarity method  & Decision Fusion & Semantic textual similarity\\
2020 & \citet{zhang2020grasp}&Grasp for stacking via deep reinforcement learning  & Stacking & Robotic arm control \\
2020 & \citet{zhang2020snapshot} &Snapshot boosting: a fast ensemble framework for deep neural networks & Boosting & Computer vision (CV) and the natural language processing (NLP) tasks\\
2021 & \citet{tsogbaatar2021iot} & DeL-IoT: A deep ensemble learning approach to uncover anomalies in IoT &Decision Fusion & IoT \\
2022&\citet{wen2020new}&A new ensemble convolutional neural network with diversity regularization for fault diagnosis &Snapshot ensemble
learning&Fault diagnosis\\
2022&\citet{hu2022representation}& Representation Learning Using Deep Random Vector Functional Link Networks for Clustering &Decision Fusion &Clustering\\

\hline
    \end{tabular}}
    \caption{Other applications}
    \label{tab:Other applications}
\end{table}

\section{Conclusions and future works}
\label{sec:Conclusions and future works}
In this paper, we reviewed the recent developments of  ensemble deep learning models. The theoretical background of ensemble learning has been elaborated to understand the success of ensemble learning. The various approaches ranging from traditional ones like bagging, boosting to the recent novel approaches like implicit/explicit ensembles, heterogeneous ensembles, have led to better performance of deep ensemble models. We also reviewed the applications of the deep ensemble models in different domains.

Although deep ensemble models have been applied across different domains, there are several open problems which can be explored in the future to fill the gap. Big data \cite{zhou2014big} is still a challenging problem, one can explore the benefits of deep ensemble models for learning the patterns using the techniques like implicit deep ensemble to maximize the performance in both time and generalization aspects. 

Deep learning models are difficult to train than shallow models as large number of weights corresponding to different layers need to be tuned. Creating deep ensemble models may further complicate the problem. Hence, randomized models can be explored to overcome the training cost.
Bagging based deep ensemble may incur heavy training time for optimizing the ensemble models. Hence, one can investigate the alternate ways of inducing diversity in the base models with lesser training cost.  
Randomized learning modules like random vector functional link network \cite{pao1994learning} are best suited for creating the ensemble models as randomized models lead to a significant variance reduction. Also, the hidden layers are randomly initialized, hence, can be used to create deep ensembles without incurring any additional cost of training \cite{katuwal2019random}. Randomized modules can be further explored using different techniques like implicit / explicit ensembles \cite{katuwal2019random}, stacking based ensembles \cite{katuwal2019stacked}. However, there are still open directions which can be worked upon like negative correlation learning, heterogeneous ensembles and so on. 

Implicit/explicit ensembles are faster compared to training of multiple deep models. However, creating diversity within a single model is a big challenge. One can explore the methods to induce more diversity among the learners within these ensembles like branching based deep models \cite{han2017branchout}. Investigate the extension of explicit/implicit ensembles to traditional models. 

Following the stacking based approach, Deep convex net (DCN) \cite{Deng2011}, traditional methods like random forest \cite{breiman2001random, zhou2017deep}, support vector machines \cite{Wang2019a, Wang2019b, Li2019} have been extended to deep learning architectures which resulted in  improved performance. One can investigate these traditional models for creating the deep ensemble models.

Another big challenge of ensemble deep learning lies in model selection for building the ensemble architecture, homogeneous and heterogeneous ensembles represent two different ways for choosing the model. However, to answer how many different algorithms, and  how many base learners in the ensemble architecture, are still problem-dependent. Finding a criterion for model selection in ensemble deep learning should be an important target for researchers in the next few years. Since most of the models focus on developing the architectures with little attention towards how to combine the base learners prediction is still unanswered. Hence, one can investigate the effect of different fusion strategies on the prediction of an ensemble output. 

For unsupervised ensemble learning or consensus clustering, the ensemble approaches include but are not limited to: Hyper-graph partitioning, Voting approach, Mutual information, etc. Consensus clustering is a powerful tool and it can improve performance in most cases. However, there are many concerns remain to be tackled, it is exquisitely sensitive, which might assert as an apparent structure without obvious demarcation or declared cluster stable without cluster resistance. Besides, current method cannot handle some complex but possible scenarios, such as the boundary samples are assigned to the single cluster, clusters do not intersect and the methods are not able to represent outliers. These are the possible research directions  for future work.

The problem of semi-supervised ensemble domains has not been extensively studied yet, and most of the literature shows that semi-supervised ensemble methods are mainly used in cases where there is insufficient labeling data. Also, combining the semi-supervision with some other machine learning methods, such as active learning, is a direction for future research.

Reinforcement learning is another popular topic recently. The idea of integrating model-based reinforcement learning with ensemble learning has been used with promising  results in many applications, but there is little integration of planning \& learning-based reinforcement learning with ensemble learning methods.

\section*{Acknowledgment}
The funding for this work is provided by the National Supercomputing Mission under DST and Miety,  Govt. of India under Grant No. DST/NSM/ R\&D\_HPC\_Appl/2021/03.29, as well as the Department of Science and Technology under Interdisciplinary Cyber Physical Systems (ICPS) Scheme grant no. DST/ICPS/CPS-Individual/2018/276. Mr. Ashwani Kumar Malik acknowledges the financial support (File no - 09/1022 (0075)/2019-EMR-I) given as scholarship by Council of Scientific and Industrial Research (CSIR),
New Delhi, India. We are  grateful to  IIT Indore for the facilities and support being provided.

\bibliographystyle{elsarticle-num-names}

\bibliography{refs,references_Mendeley}
\end{document}